\documentclass[letterpaper]{article} 
\usepackage{aaai2026}  
\usepackage{times}  
\usepackage{helvet}  
\usepackage{courier}  
\usepackage[hyphens]{url}  
\usepackage{graphicx} 
\usepackage{natbib}  
\usepackage{caption} 
\usepackage{algorithm}
\usepackage{algorithmic}

\usepackage{newfloat}
\usepackage{listings}
\DeclareCaptionStyle{ruled}{labelfont=normalfont,labelsep=colon,strut=off} 
\floatstyle{ruled}
\newfloat{listing}{tb}{lst}{}
\floatname{listing}{Listing}
\usepackage{amsmath}
\usepackage{booktabs}

\title{Causal-HalBench: Uncovering LVLMs Object Hallucinations Through Causal Intervention}
\author {
    Zhe Xu \equalcontrib \textsuperscript{\rm 1},
    Zhicai Wang \equalcontrib \textsuperscript{\rm 1},
    Junkang Wu\textsuperscript{\rm 1},
    Jinda Lu \textsuperscript{\rm 1},
    Xiang Wang \textsuperscript{\rm 1} \thanks{Corresponding author}
}
\affiliations {
    \textsuperscript{\rm 1}University of Science and Technology of China\\
    \{xz2002,wangzhic,jkwu0909,lujd\}@mail.ustc.edu.cn, xiangwang1223@gmail.com,
}

\usepackage{bibentry}

\begin{document}

\maketitle

\begin{abstract}
Large Vision-Language Models (LVLMs) often suffer from object hallucination, making erroneous judgments about the presence of objects in images. We propose this primarily stems from spurious correlations arising when models strongly associate highly co-occurring objects during training, leading to hallucinated objects influenced by visual context. Current benchmarks mainly focus on hallucination detection but lack a formal characterization and quantitative evaluation of spurious correlations in LVLMs. To address this, we introduce causal analysis into the object recognition scenario of LVLMs, establishing a Structural Causal Model (SCM). Utilizing the language of causality, we formally define spurious correlations arising from co-occurrence bias. To quantify the influence induced by these spurious correlations, we develop Causal-HalBench, a benchmark specifically constructed with counterfactual samples and integrated with comprehensive causal metrics designed to assess model robustness against spurious correlations. Concurrently, we propose an extensible pipeline for the construction of these counterfactual samples, leveraging the capabilities of proprietary LVLMs and Text-to-Image (T2I) models for their generation. Our evaluations on mainstream LVLMs using Causal-HalBench demonstrate these models exhibit susceptibility to spurious correlations, albeit to varying extents.
\end{abstract}

\begin{links}
    \link{Code}{https://github.com/zhexu-ustc/Causal-HalBench}
\end{links}

\section{Introduction}
\begin{figure*}[htbp]
  \centering
  \includegraphics[width=\linewidth]{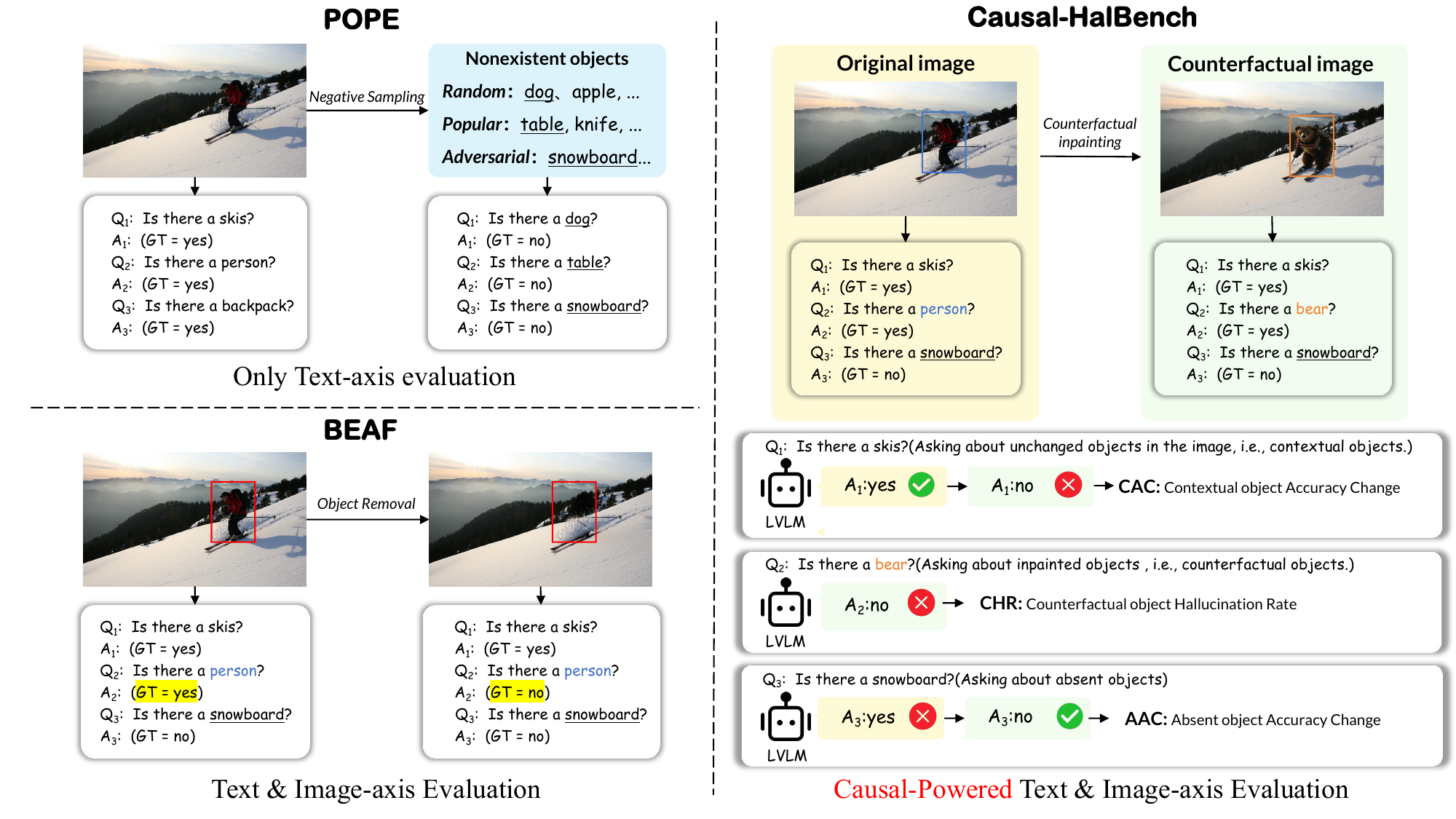}
  \caption{\textbf{Illustration of Causal-HalBench.} We present a comparison between current hallucination benchmarks and our Causal-HalBench for Object Hallucination in LVLMs.}
  \label{fig:causal_halbench_intro}
\end{figure*}

Recent advancements in Large Language Models (LLMs) \cite{chiang2023vicuna,touvron2023llama} have driven the emergence of Large Vision-Language Models (LVLMs) \cite{liu2023visual,dai2023instructblip,hurst2024gpt}. By integrating a visual encoder with an LLM framework, these models have significantly extended LLMs' capabilities into the visual domain. Consequently, object recognition, a core task in computer vision, has become a crucial metric for evaluating LVLMs' performance. Typically, assessing object recognition in LVLMs involves querying the model about an object's presence, to which it usually responds with a ``yes" or ``no" \cite{li2023evaluating,ye2024beaf}. However, many studies report that LVLMs frequently suffer from object hallucination, a phenomenon where the models make erroneous judgments about the presence of objects in an input image \cite{li2023evaluating,zhai2023halle,zhou2023analyzing}. Prior work has indicated that spurious correlations arising from image datasets as a significant factor impeding object recognition \cite{choi2012context,neuhaus2023spurious}. Specifically, models learn associations between highly co-occurring objects due to imbalanced object distributions within the data. However, these objects are not causally related. This leads to models making predictions based on these learned prior associations, resulting in incorrect judgments. This phenomenon is known as spurious correlations. For example, as depicted in Figure \ref{fig:causal_halbench_intro} (right), the model might fail to correctly identify skis under a bear's paw due to the absence of a human. Since LVLMs are trained on image data, these spurious correlations are also a primary cause of object hallucination. Yet, accurately quantifying and evaluating their impact remains a significant challenge.

Several efforts have been made to evaluate and understand object hallucination in LVLMs \cite{li2023evaluating,ye2024beaf,wang2023amber,chen2024multi}. We categorize these works based on their data construction strategies into two main types: Only Text-axis Evaluation and Text \& Image-axis Evaluation. Only Text-axis Evaluation methods directly construct questions from existing image data, such as POPE \cite{li2023evaluating}, NOPE \cite{lovenia2023negative}, and CIEM \cite{hu2023ciem}. For instance, POPE (Figure \ref{fig:causal_halbench_intro}, top-left corner) uses a discriminative question-answering format: ``Is there \{object\} in this image?" and constructs positive and negative questions via image annotations and various negative sampling strategies. Conversely, Text \& Image-axis Evaluation methods simultaneously construct both images and questions, including ROPE \cite{chen2024multi} and BEAF \cite{ye2024beaf}. BEAF (Figure \ref{fig:causal_halbench_intro}, bottom-left corner), for example, uses an image editing model to remove objects and then formulates related questions.

Despite their effectiveness in hallucination detection, these methods lack a formal definition of spurious correlations. For instance, BEAF solely assesses a model's perception of image changes through object removal, lacking further discussion on spurious correlations within the model. To address these gaps, we first formalize spurious correlations in LVLMs using a Structural Causal Model (SCM) \cite{pearl2009causal}. Then we propose Causal-HalBench, a causal benchmark for evaluating object hallucination in LVLMs. Inspired by causal intervention \cite{pearl2012calculus}, we leverage a Visual Content Intervention (VCI) technique to break spurious correlations via introducing counterfactual visual content. To realize VCI, we develop an automated, scalable data synthesis pipeline, which utilizes T2I models to replace high co-occurrence objects with low co-occurrence objects. As illustrated in Figure \ref{fig:causal_halbench_intro} (right), we construct 1387 counterfactual images from 757 original images via counterfactual inpainting, yielding 9709 corresponding Question-Answering (QA) pairs. We also propose novel causal-based metrics to quantify the impact of spurious correlations by observing model responses to images before and after changes. Our experiments on mainstream LVLMs reveal that spurious correlations are widespread in LVLMs, leading to hallucination across various aspects. Furthermore, we find that newer models may be more susceptible to spurious correlations.

In summary, our main contributions are:

\begin{itemize}
\item  We are the first to integrate causal analysis into LVLM object hallucination study, systematically analyzing and quantifying spurious correlation influence using SCMs.
\item  We develop an automated and scalable pipeline leveraging T2I models and proprietary LVLMs for high-quality causal counterfactual sample generation.
\item  We establish Causal-HalBench, the first causal hallucination detection benchmark, comprising over 10,00 counterfactual samples and comprehensive evaluation metrics. Through extensive experiments on Causal-HalBench, we provide empirical evidence of state-of-the-art LVLMs' susceptibility to spurious correlations, identifying a key area for improving model faithfulness.
\end{itemize}

\section{Related Work}

\noindent{\textbf{Large Vision-Language Models.}} The past few years have witnessed rapid advancements in Large Vision-Language Models (LVLMs), which integrate powerful visual encoders with large language model (LLM) architectures to achieve multimodal understanding and generation capabilities. Early pioneers like CLIP \cite{radford2021learning} and ALIGN \cite{jia2021scaling} demonstrate robust vision-language alignment. More recent advancements, such as LLaVA \cite{liu2023visual}, Mplug-Owl\cite{ye2023mplug} and GPT-4o \cite{hurst2024gpt} have pushed the boundaries, exhibiting remarkable zero-shot and few-shot performance in various visual tasks, such as visual question answering \cite{hudson2019gqa} and image captioning \cite{agrawal2019nocaps}.

\noindent{\textbf{Object Hallucination in LVLMs.}} Despite impressive capabilities, LVLMs often generate objects inconsistent with images, a problem known as object hallucination \cite{li2023evaluating,zhai2023halle}. Several methods have been suggested to mitigate the object hallucination issue \cite{zhai2023halle,gunjal2024detecting}. To quantify progress on mitigating them, various benchmarks have been developed \cite{rohrbach2018object,li2023evaluating,ye2024beaf,wang2023amber,chen2024multi}. While effective for detection, these benchmarks lack a causal analysis framework to quantify the causal effect strength of specific spurious correlations on predictions. This limits a deeper, causal understanding of hallucination, which our work addresses.

\noindent{\textbf{Spurious Correlations/Biases in Language Model.}} Spurious correlations constitute a critical issue in LVLMs: models often learn misleading statistical associations between co-occurring objects in training data (without true causal relationships), impairing their correct prediction ability and significantly contributing to problems such as object hallucination. While causal analysis has emerged as a powerful tool for addressing analogous biases in Natural Language Processing (NLP) \cite{wang2022should,zhu2022generalizing,wang2023causal,wang2023fragile}, this approach remains underutilized for spurious correlation-induced hallucination in LVLM. Although benchmarks like MM-SPUBENCH \cite{ye2024mm} and other existing methods \cite{li2023evaluating,ye2024beaf,wang2023amber} explore spurious biases, they primarily focus on detecting bias, yet lack the capacity for rigorous causal analysis and examination of these spurious biases\cite{zhu2024boosting,zhu2024enhancing}. Our work directly addresses this critical gap by explicitly applying a causal framework to examine and quantify the causal impact of spurious correlations on object hallucination in LVLMs.

\section{Methodology}

In this section, we first leverage SCM to formalize spurious correlations of object hallucination in LVLMs (\S3.1). Then we introduce Visual Content Intervention (VCI), a curated causal intervention technique that breaks the non-causal path by manipulating the visual content of images. This enables us to quantify their causal effect using Average Causal Effect (ACE) and Direct Causal Effect (DCS) (\S3.2).

\subsection{Causal Analysis of Object Hallucination}

\begin{figure}[h]
  \centering  \includegraphics[width=\linewidth]{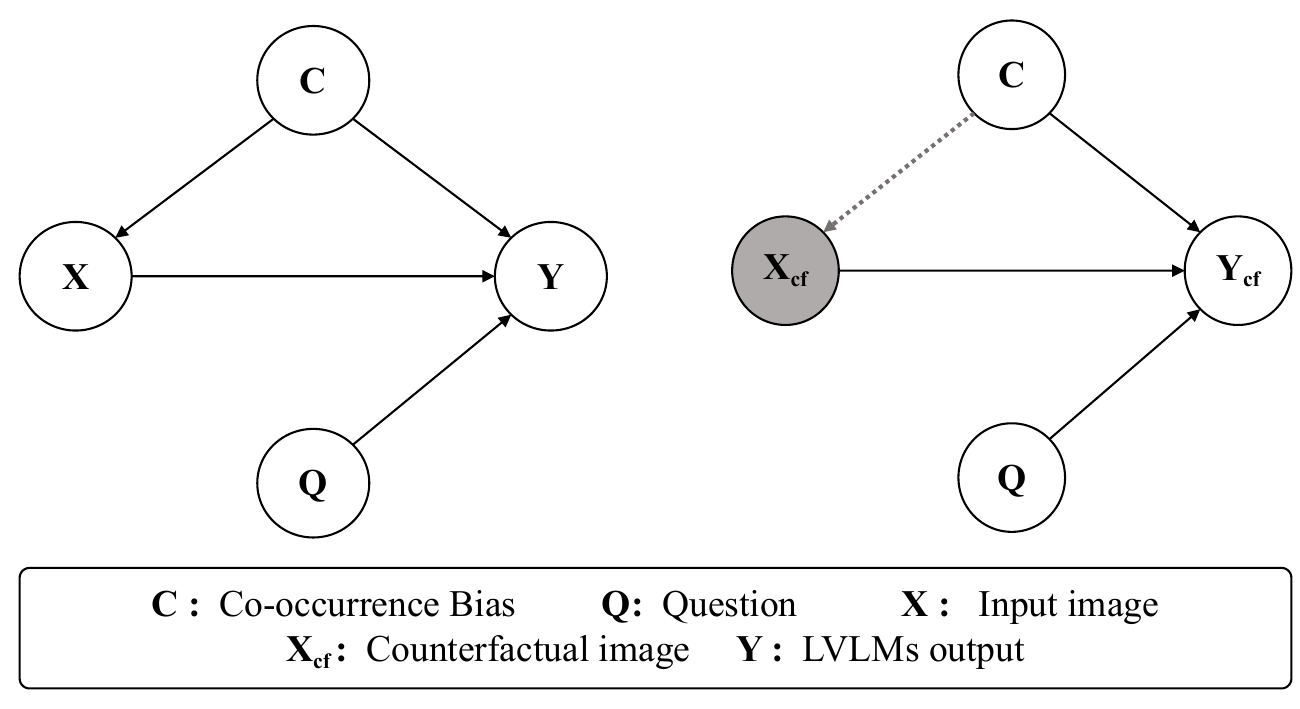}
  \caption{\textbf{Illustration of SCM.} We present the original SCM of spurious correlation (left), contrasted with the SCM derived via VCI (right).}
  \label{fig:scm}
\end{figure}

To analyze how spurious correlations cause object hallucination in LVLMs, we propose a SCM, a powerful tool that leverages structural theory to estimate causal effects from data \cite{pearl2000models}, as shown in Figure \ref{fig:scm} (left). In this SCM, $X$ represents the input image, $Q$ denotes the question provided to the LVLM (e.g., ``Is there \{object\} in the image?''), and $Y$ signifies the LVLM's output regarding object existence. A key variable is 
C, referred to as \textbf{Co-occurrence Bias}. This bias denotes the statistical patterns (from training data) of frequent co-occurrence of specific objects/concepts, reflecting the model’s tendency to associate these typically non-causal learned relationships. Ideally, an LVLM should predict accurately based only on $X$ (the direct causal path $X \rightarrow Y$), with $Q$ directly influencing $Y$ via $Q \rightarrow Y$.

However, as Figure \ref{fig:scm} (left) illustrates, $C$ functions as a confounder in the relationship between $X$ and $Y$  \cite{pearl2009causal}-a variable that influences both the input and the output, causing a spurious correlation between them. In our SCM, the link $C \rightarrow X$ shows that input images ($X$), particularly those in the training data, inherently contain co-occurrence patterns driven by this Co-occurrence Bias ($C$); while the path $C \rightarrow Y$ indicates that the model directly uses this Co-occurrence Bias ($C$) to generate its output ($Y$), serving as a cognitive shortcut that bypasses in-depth analysis of $X$. This confounding effect creates a backdoor path: $X \leftarrow C \rightarrow Y$. Spurious correlations form through this path. When an LVLM over-relies on this pathway, it may predominantly leverage Co-occurrence Bias ($C$) embedded in $X$ rather than the visual information within $X$ to predict the presence of an object ($Y$), thereby leading to hallucinations.

\subsection{Visual Content Intervention}

In the causal analysis presented in the previous section, we treat $C$ as a confounder between $X$ and $Y$. Prior research has extensively explored confounder analysis \cite{keith2020text,qian2021counterfactual,feder2022causal,weld2022adjusting}, and central to these methods is the concept of intervention (denoted as $\text{do}(\cdot)$) \cite{pearl2012calculus}. Intervention refers to the manipulation of a specific node (variable) within a causal system, which effectively wipes out all its incoming causal links and sets it to a specific value, thereby allowing us to isolate its true, unconfounded effects.

We propose \textbf{Visual Content Intervention (VCI)} as our tailored method for spurious correlations. The causal mechanism of VCI is depicted in Figure \ref{fig:scm} (right). Specifically, VCI systematically constructs a counterfactual image ($x_{cf}$) from an original image by introducing counterfactual visual content. This strategic modification of $X$ disrupts the influence of $C$ inherent in the original image, by altering its associated visual cues. The formal expression for the outcome under this intervention is:

\begin{align}
y_{cf} &= Y(\text{do}(X=x_{cf})).
\end{align}
The comparison between the intervened and original outcomes , central to counterfactual analysis  \cite{pearl2019causal}, isolates the causal effect of spurious correlations. The average of these isolated effects defines the \textbf{Average Causal Effect (ACE)} \cite{rubin1974estimating}, a
fundamental causal inference concept, We formally define ACE as follows:

\begin{align}
ACE &= E[Y \mid \text{do}(X=X_{cf}), Q] - E[Y \mid X, Q].
\end{align}
Beyond ACE, we also introduce the \textbf{Direct Causal Strength (DCS)}. DCS quantifies the model's reliance on actual visual information under controlled spurious correlations. It is formally represented by the expected outcome under intervention:

\begin{align}
DCS &= E[Y \mid \text{do}(X=x_{cf}), Q].
\end{align}
ACE and DCS form the fundamental causal basis for quantifying the effects of spurious correlations. By introducing specific definitions, we operationalize these concepts into measurable metrics for our benchmark. These will be detailed in the subsequent section.

\section{Causal-HalBench}

Based on the VCI proposed in the preceding section, we utilize inpainting models and LVLMs to develop an automated, scalable pipeline for high-quality counterfactual sample generation (\S4.1). Subsequently, based on our proposed pipeline, we establish Causal-HalBench and provide an overview of our dataset (\S4.2). Finally, we construct comprehensive metrics based on our prior causal analysis to quantify the effect of spurious correlation in LVLMs (\S4.3).

\begin{figure*}[h]
  \centering
  \includegraphics[width=\linewidth]{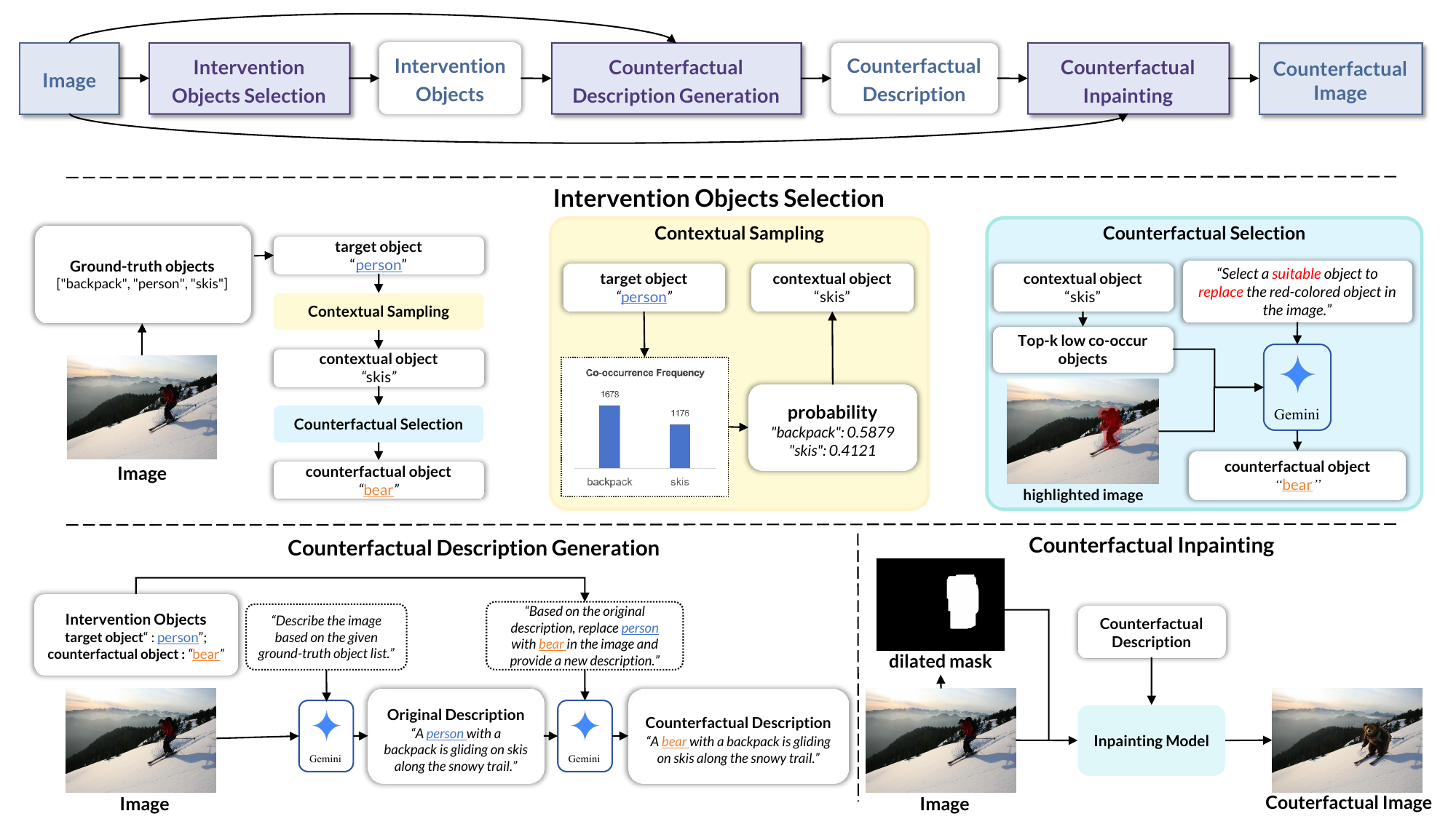}
  \caption{\textbf{The Data Construction Pipeline of Causal-HalBench.} We propose a three-stage, fully automated, and scalable pipeline for constructing counterfactual samples in Causal-HalBench.}
  \label{fig:pipeline}
\end{figure*}

\subsection{Dataset Construction Pipeline}
To implement Visual Content Intervention (VCI) and generate counterfactual samples ($X_{cf}$) for causal analysis, we design an automated and scalable pipeline. This three-stage pipeline, illustrated in Figure \ref{fig:pipeline}, systematically creates counterfactual samples through: 1) Intervention Objects Selection, 2) Counterfactual Description Generation, and 3) Counterfactual Inpainting. We offer a detailed explanation of each stage as follows. For more details, please refer to the Appendix B.1.

\noindent{\textbf{Intervention Objects Selection.}}     
The initial stage of our pipeline is critical for identifying intervention objects that fulfill two key requirements for effective counterfactual sample generation: ensuring low co-occurrence with the original context and enhancing inpainting success. This process begins by designating each annotated object in the original image as a target object $o_t$. For every $o_t$, we employ \textbf{Contextual Sampling} to probabilistically sample a contextual object $o_c$ based on its co-occurrence frequency with $o_t$ , ensuring visual relevance and diversity. Subsequently, for \textbf{Counterfactual Selection}, we form a candidate pool of top-k objects with the lowest co-occurrence frequencies with $o_c$. Given the original image with the highlighted target object, the powerful proprietary LVLM Gemini \cite{team2023gemini} then selects the most appropriate object from this candidate pool to replace $o_t$, prioritizing visual suitability to ensure the resulting counterfactual object $o_{cf}$ is not only a low-association adversarial visual element but also highly amenable to inpainting, significantly increasing the inpainting success rate. This meticulous approach ultimately yields ($o_t$, $o_c$, $o_{cf}$) triplets, forming the indispensable basis for subsequent counterfactual sample generation.

\noindent{\textbf{Counterfactual Description Generation.}}
The second stage of our pipeline aims to create accurate, high-quality counterfactual descriptions that will serve as inpainting prompts. These descriptions are crafted with the help of Gemini. Specifically, Gemini is first prompted to generate a description of the original image based on its ground-truth object annotations. Subsequently, leveraging this initial description, Gemini is further prompted to incorporate the conceptual replacement of the target object $o_t$ with the previously selected counterfactual object $o_{cf}$, thereby constructing a reasonable and accurate counterfactual description. This ensures the generated description accurately reflects the desired visual modification for the subsequent inpainting process.

\noindent{\textbf{Counterfactual Inpainting.}}
The final stage of our pipeline is dedicated to generating high-quality counterfactual samples ($X_{cf}$). For this purpose, we employ FLUX-controlnet inpainting \cite{zhang2023adding,flux2024} as our core inpainting model. We first utilize the Segment Anything Model (SAM) \cite{kirillov2023segment} to extract precise masks for the objects. Then, to ensure comprehensive coverage and minimize the potential influence of specific object shapes, these masks are subsequently dilated. Finally, the original image, the dilated mask, and the counterfactual description are input into the inpainting model, yielding the desired counterfactual image.

\noindent{\textbf{Discussion.}} We acknowledge that using an inpainting model is an approximation of the theoretical $do(X=x_{cf})$ intervention. Potential visual artifacts introduced by the inpainting process could act as interfering factors, such as unnatural foreground-background integration and the disruption of context objects. To mitigate this, we employed several strategies: (1) using a state-of-the-art inpainting model (FLUX-controlnet) to ensure high fidelity; (2) performing manual filtering to discard low-quality images; and (3) as we will show in Section 5.3, we use CLIP Score to quantitatively verify that the synthetic region correctly represents the counterfactual object while diminishing the signal of the original target object \cite{ye2024beaf}, thus providing evidence for the effectiveness of our intervention approximation.

\begin{table}[tp]
    \small
    \centering
    \begin{tabular}{lcccc}
        \toprule
        \textbf{} & \textbf{Image} & \textbf{Img-Q Pair} & \textbf{Yes} & \textbf{No} \\
        \midrule
        Original & 757 & 4161 & 1387 & 2774 \\
        Counterfactual & 1387 & 5548 & 2774 & 2774 \\
        \midrule
        Total & 2144 & 9709 & 4161 & 5548 \\
        \bottomrule
    \end{tabular}
    \caption{\textbf{Dataset Statistics.} Causal-HalBench contains 9.7K image-question pairs, consisting of the original and counterfactual ones.}
    \label{tab:Statistics}
\end{table}

\subsection{Dataset Overview}

Our dataset is made up of images and their corresponding Q\&A pairs. The images are split into original and counterfactual categories. We first randomly sample original images from the MSCOCO dataset's validation set \cite{lin2014microsoft} and then use our designed pipeline to generate corresponding counterfactual images. After a round of manual filtering, we finally obtain a total of 757 original images and 1387 generated counterfactual images. We adopt the question format ``Is there \{object\} in this image?", consistent with POPE\cite{li2023evaluating}. For positive questions where the answer is ``yes", we query the contextual object $o_c$ for both original and counterfactual images, and the counterfactual object $o_{cf}$ specifically for counterfactual images. For negative questions where the answer is ``no", we query the top-$k$ absent objects ($o_a$) that frequently appear with the original target object ($o_t$). Ultimately, our dataset includes 2144 images and 9709 image-question pairs; see Table \ref{tab:Statistics} for specific statistical information.

\subsection{Evaluation Metrics}
Building upon the concepts of ACE and DCS, we design our evaluation metrics. We first establish three question types: \textbf{contextual questions ($Q_c$)} for a contextual object $o_c$ (GT: Yes); \textbf{counterfactual questions ($Q_{cf}$}) for a counterfactual object $o_{cf}$ (GT: Yes); and \textbf{absent questions ($Q_a$)} for an object $o_a$ (GT: No). We then formalize the model's output, $Y$, as a binary variable, where $Y$=1 for a correct answer and $Y$=0 for an incorrect one. This allows the expectation, $E[Y]$, within the ACE and DCS to be directly interpreted as the model's accuracy. We define the model's overall accuracy on an image set $X$ and a question set $Q$ as $Acc(X,Q)$. Based on these definitions, we formulate our metrics as follows.

\noindent{\textbf{Contextual object Accuracy Change (CAC).}}
CAC quantifies the decrease in accuracy of $Q_c$ from the original image to the counterfactual image, which represents ACE for $Q=Q_c$. 
\begin{align}
CAC = Acc(X, Q_c) - Acc({X}_{cf}, Q_c).
\end{align}
A higher CAC reflects a stronger causal effect of spurious correlations, indicating that the model is more prone to making predictions based on co-occurrence biases.

\noindent{\textbf{Absent object Accuracy Change (AAC).}}
AAC quantifies the increase in accuracy of $Q_a$ from the original image to the counterfactual image, which represents ACE for $Q=Q_a$.
\begin{align}
AAC = Acc({X}_{cf}, Q_a) - Acc(X, Q_a).
\end{align}
A higher AAC indicates that the model is more influenced by spurious correlations, leading it to hallucinate related non-existent objects.

\noindent{\textbf{Counterfactual object Hallucination Rate (CHR).}}
CHR measures the hallucination rate for the counterfactual object on counterfactual images, which reflects DCS, with $Acc({X}_{cf}, Q_{cf})$ corresponding to DCS for $Q=Q_{cf}$.
\begin{align}
CHR = 1 - Acc({X}_{cf}, Q_{cf}).
\end{align}
A lower CHR reflects the model's strong visual perception and weaker influence from spurious correlations. 

\section{Experiment}

\begin{table*}[htbp]
    \small
    \centering
    \begin{tabular}{lcccc|cccc}
        \toprule
        & \multicolumn{2}{c}{\textbf{Original Image}} & \multicolumn{2}{c}{\textbf{Counterfactual Image}} & \multicolumn{4}{c}{\textbf{Metrics}} \\
        \cmidrule(lr){2-3} \cmidrule(lr){4-5} \cmidrule(lr){6-9}
        \textbf{Model} & \textbf{Acc(Q$_{c}$)$\uparrow$} & \textbf{Acc(Q$_{a}$)$\uparrow$} & \textbf{Acc(Q$_{c}$)$\uparrow$} & \textbf{Acc(Q$_{a}$)$\uparrow$} & \textbf{Acc$\uparrow$} & \textbf{$\Delta$Acc(Q$_{c}$)$\downarrow$} & \textbf{$\Delta$Acc(Q$_{a}$)$\downarrow$} & \textbf{CHR$\downarrow$} \\
        \midrule
        LLaVA-NEXT-8B & 90.9 & 81.1 & 86.4 & 82.2 & 85.9 & 4.5 & 1.1 & 6.8 \\
        LLaVA-onevision-7B & 82.8 & 94.8 & 79.2 & 95.0 & 87.0 & 3.6 & 0.2 & 14.4 \\
        Kimi-VL-A3B & 91.0 & 75.2 & 85.6 & 85.4 & 84.4 & 5.4 & 10.2 & 7.4 \\
        MiniCPM-o-2\_6 & 89.8 & 78.1 & 86.6 & 80.9 & 83.1 & 3.2 & 2.8 & 14.9 \\
        InternVL2.5-8B & 88.9 & 72.4 & 86.1 & 79.5 & 78.8 & 2.8 & 7.1 & 29.3 \\
        mPLUG-Owl3-7B & 81.8 & 93.7 & 77.9 & 94.0 & 86.6 & 3.8 & 0.3 & 14.4 \\
        Qwen2.5-VL-7B & 70.0 & 97.6 & 68.2 & 98.1 & 84.1 & 1.8 & 0.5 & 27.3 \\
        \midrule
        GPT-4o & 79.1 & 91.0 & 75.5 & 92.5 & 84.3 & 3.6 & 1.5 & 12.4 \\
        Gemini1.5-pro & 81.9 & 87.4 & 73.8 & 87.7 & 86.1 & 8.1 & 0.3 & 21.4 \\
        \bottomrule
    \end{tabular}
    \caption{\textbf{Evaluation Results on Causal-HalBench.} We evaluate various LVLMs on Causal-HalBench. Acc(Q$_{c}$) and Acc(Q$_{a}$) report the model’s accuracy on contextual objects and absent objects in the original and counterfactual images, while $\Delta$Acc(Q$_{c}$) and $\Delta$Acc(Q$_{a}$) indicate the resulting changes. All values in the table are percentages. Underlined values indicate the best results.}
    \label{tab:Evaluation results}
\end{table*}

In this section, we conduct a series of experiments on mainstream LVLMs leveraging Causal-HalBench. First, we evaluate various LVLMs on Causal-HalBench using our proposed causality-based metrics and analyze the experimental results (\S5.1). Subsequently, we compare our Causal-HalBench with representative object hallucination benchmarks, encompassing both discriminative (POPE \cite{li2023evaluating}) and generative (CHAIR \cite{rohrbach2018object}) approaches (\S 5.2). Finally, we perform additional study on Causal-HalBench, visualizing its co-occurrence patterns and assessing the dataset quality utilizing CLIP Score \cite{hessel2021clipscore} (\S5.3).

\noindent{\textbf{Implementation Details.}} All the reported performance is the zero-shot performance of LVLMs on a single A40, including GPT-4o \cite{hurst2024gpt}, Gemini1.5-pro \cite{team2024gemini}, LLAvA-NEXT \cite{liu2024llavanext}, LLAvA-onevision \cite{li2024llava}, Kimi-VL \cite{team2025kimi}, MiniCPM \cite{team2025minicpm}, InternVL2.5 \cite{chen2024expanding}, mPLUG-Owl3 \cite{ye2024mplug}, and Qwen2.5-VL \cite{bai2025qwen2}. For more details, please refer to the Appendix B.2.

\subsection{Evaluation Results on Causal-HalBench}
We evaluate several LVLMs on Causal-HalBench and analyze the results. Experimental results are shown in Table \ref{tab:Evaluation results}. 

\noindent{\textbf{Main Results.}} As presented in Table \ref{tab:Evaluation results}, the majority of models demonstrate high CAC values. Notably, Qwen2.5-VL-7B achieve the lowest CAC at 1.8\%, yet this desirable outcome is accompanied by a diminished Acc(Q$_c$) compared to other models. Conversely, in the evaluation of AAC, Kimi-VL-A3B record the highest AAC value at 10.2\%, substantially exceeding all other models; InternVL2.5-8B (7.1\%) and MiniCPM-o-2\_6 (2.8\%) followed, while the remaining models demonstrate lower AAC values. Finally, regarding CHR, LLaVA-NEXT-8B exhibit the optimal performance at 6.8\%, with Kimi-VL-A3B also demonstrating commendable results at 7.4\%. In contrast, InternVL2.5-8B (29.3\%) and Qwen2.5-VL-7B (27.3\%) perform notably worse in this regard. Meanwhile, the overall performance of closed-source models do not significantly outperform open-source models. 

\noindent{\textbf{Analysis.}} Our experimental results indicate that LVLMs are susceptible to spurious correlations to varying extents, with nearly all models exhibiting high CAC values, confirming their general vulnerability to spurious correlations when contextual elements are altered. We also observe that the influence of spurious correlations on models is multifaceted. For instance, while Kimi-VL-A3B demonstrates exceptional performance in CHR, it exhibits the poorest performance in AAC. Conversely, Qwen2.5-VL-7B performs favorably on AAC but shows noticeably inferior results in CHR . This underscores the inherent limitations of relying on a single accuracy metric in traditional evaluation methodologies, thereby necessitating a comprehensive evaluation across multiple metrics to thoroughly assess spurious correlations within these models. Furthermore, the aforementioned findings suggest that while more recent models (e.g., Qwen2.5-VL) may outperform earlier iterations (e.g., LLaVA-NEXT) on certain general benchmarks, their susceptibility to spurious correlations can be markedly more pronounced. This phenomenon can be attributed to the increased scale of training data, thereby underscoring the critical importance of detecting and mitigating spurious correlations for the continued advancement of LVLMs.

\subsection{Comparison with Other Benchmarks}

\noindent To validate the efficacy of our benchmark, we conduct a comparative analysis of Causal-HalBench against representative object hallucination benchmarks, specifically including both discriminative (POPE\cite{li2023evaluating}) and generative (CHAIR\cite{rohrbach2018object})) approaches.

\noindent{\textbf{Comparison with POPE.}} In the Table \ref{tab:POPE Comparison}, we evaluate model performance on both POPE and the entire Causal-HalBench, using accuracy as the primary evaluation metric. A discernible consistency in performance trends emerge between the two benchmarks. Critically, however, the deliberate inclusion of counterfactual samples in Causal-HalBench leads to a subdued performance compared to POPE. This finding affirms Causal-HalBench's robustness for conventional evaluation, while simultaneously underscoring its superior extensibility beyond POPE, particularly through its integrated, causality-based metrics designed for a precise quantification of spurious correlations.

\begin{table}[htbp]
    \small
    \centering
    \begin{tabular}{lcc} 
        \toprule
        \textbf{Model} & \textbf{Causal-HalBench} & \textbf{POPE} \\
        \midrule
        LLaVA-NEXT-8B & 85.9 & 87.1 \\
        LLaVA-onevision-7B & 87.0 & 88.4 \\
        Kimi-VL-A3B & 84.4 & 88.5 \\
        MiniCPM-o-2\_6 & 83.1 & 86.7 \\
        InternVL2.5-8B & 78.8 & 88.7 \\
        mPLUG-Owl3-7B & 86.6 & 87.1 \\
        Qwen2.5-VL-7B & 84.1 & 85.9 \\
        \bottomrule
    \end{tabular}
    \caption{\textbf{Comparison of Performance on POPE and Causal-HalBench.} We compare model accuracy on POPE and the entire Causal-HalBench.}
    \label{tab:POPE Comparison}
\end{table}

\begin{table}[htbp]
    \small
    \centering
    \begin{tabular}{lcccc}
        \toprule
        \textbf{Model} & \textbf{CHAIR-S$\downarrow$} & \textbf{CHAIR-I$\downarrow$} & \textbf{CHR$\downarrow$} \\
        \midrule
        Qwen2.5-VL-7B & 44.1 (54.7) & 15.8 (20.0) & 27.3 \\
        InternVL2.5-8B & 68.2 (76.2) & 18.7 (22.7) & 29.3\\
        MiniCPM-o-2\_6 & 22.9 (30.0) & 13.7 (17.8) & 14.9 \\
        LLaVA-onevision-7B & 19.3 (26.9) & 11.1 (15.2) & 14.4\\
        mPLUG-Owl3-7B & 24.5 (33.6) & 14.3 (18.4) & 14.4\\
        Kimi-VL-A3B & 34.9 (43.1) & 14.6 (18.1) & 7.4 \\
        LLaVA-NEXT-8B & 29.0 (38.8) & 13.7 (18.1) & 6.8 \\
        \bottomrule
    \end{tabular}
    \caption{\textbf{Comparison of CHAIR metrics and CHR on Causal-HalBench.} We measure the CHAIR score based on the answers derived from the images in BEAF using the prompt “Provide a brief description of the given image.” Note: Results in parentheses are measured on a subset containing only counterfactual images within Causal-HalBench.}
    \label{tab:CHAIR Comparison}
\end{table}

\noindent{\textbf{Comparison with CHAIR.}} Beyond discriminative testing, we also conduct open generation testing on Causal-HalBench. For each image, we prompt the model with ``Provide a brief description of the given image." In Tables \ref{tab:CHAIR Comparison}, we present the CHAIR results for both the full dataset and a subset containing only counterfactual images, alongside the CHR results. CHAIR is a popular metric for evaluating object hallucination in image captioning tasks. It comprises CHAIR-I and CHAIR-S, which respectively count the hallucinated instances and sentences containing these objects in the generated output. The lower the score, the better. From the results, we observe that if the model includes more hallucinated objects and sentences in its generated output, then CHR tends to be higher, which demonstrates the effectiveness of our evaluation method. Simultaneously, the model's hallucination problem worsened during open generation for counterfactual images, further illustrating the effect of spurious correlations on model hallucinations.

\subsection{Additional Study on Causal-HalBench}

\begin{figure}[h]
  \centering  \includegraphics[width=\linewidth]{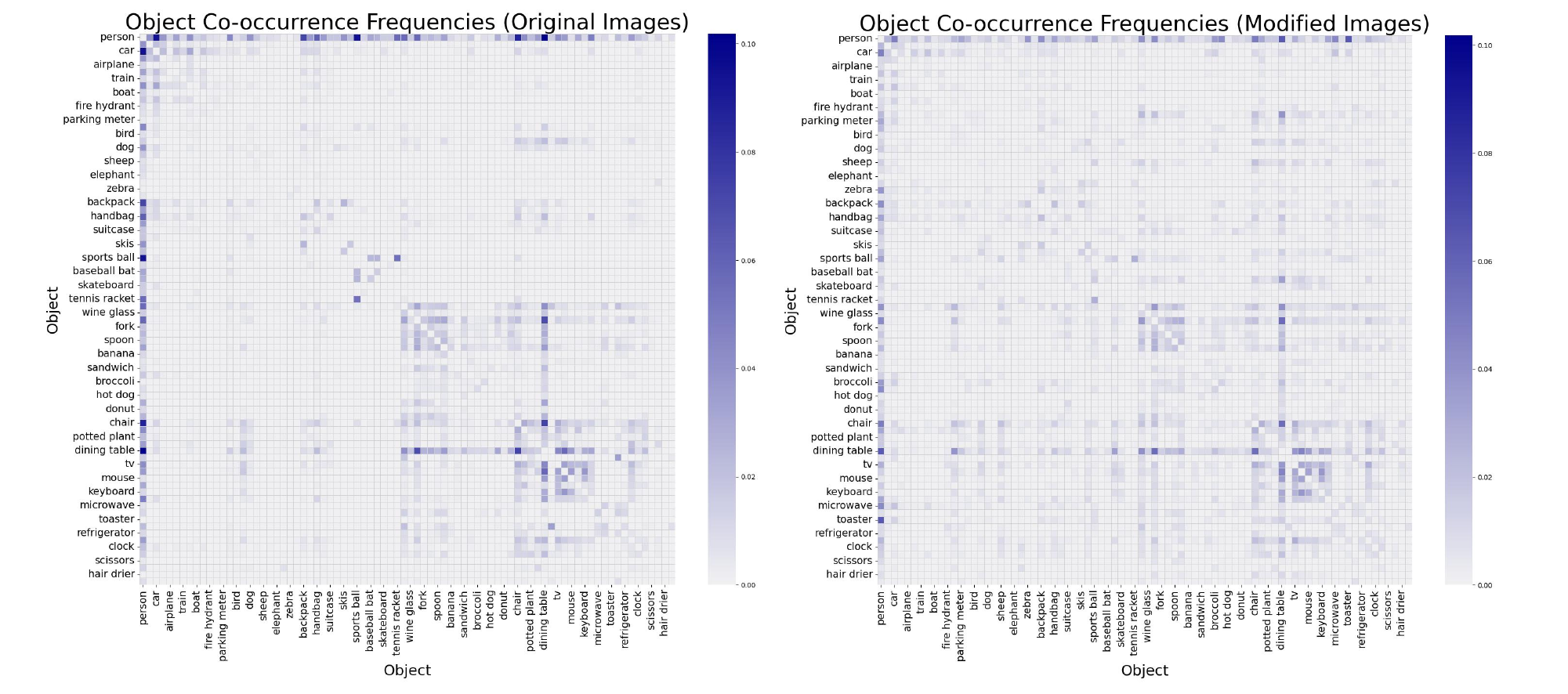}
  \caption{\textbf{Visualization of Co-occurrence Patterns.} We present the heatmap of the co-occurrence matrix for original images (left), contrasted with the heatmap of the co-occurrence matrix for modified images (right).}
  \label{fig:heatmap}
\end{figure}

\noindent{\textbf{Balanced Co-occurrence Patterns.}} To systematically evaluate the effects of spurious correlations within LVLMs, our proposed Causal-HalBench intentionally enriches and balances co-occurrence patterns in image data through VCI. To visualize this comparison, we calculate the co-occurrence matrices between objects in both the original and modified image datasets and plot their respective heatmaps, as shown in Figure \ref{fig:heatmap}. We clearly observe a distinct shift: compared to the original image dataset, which exhibits significantly concentrated high-frequency co-occurrence patterns between certain objects (manifesting as darker regions), the modified image dataset demonstrably increases the frequency of previously low-frequency or even non-co-occurring object pairs (indicated by a greater variety of shaded regions rather than uniform white). Concurrently, this process effectively dilutes the strength of pre-existing strong correlations, ultimately presenting a more balanced distribution of co-occurrence patterns.

\noindent{\textbf{Synthetic Data Quality Evaluation.}}  We manually filter the dataset to ensure the quality of the synthesized data. However, considering that most current LVLMs adopt CLIP as their visual encoder, we further evaluate the quality of the synthesized data using CLIP Score. We measure the CLIP score between an object and a prompt ``a photo of \{object\}". We crop the inpainting region of the image and set the \{object\} in the prompt to be the target object and the counterfactual object, respectively. As shown in Table \ref{tab:CLIP Score}, the synthesized images have a lower CLIP Score for the target object and a higher CLIP Score for the counterfactual object compared to the original images, which ensures the quality of the synthesized data.

\begin{table}[htbp]
    \small
    \centering
    \begin{tabular}{lcc} 
        \toprule 
        \textbf{Type} & \textbf{Target} & \textbf{Counterfactual} \\
        \midrule 
        Original    & 26.5 & 20.6 \\ 
        Synthetic   & 22.4 & 27.5 \\ 
        \bottomrule
    \end{tabular} 
    \caption{\textbf{CLIP Score Results of Causal-HalBench.} We measure the CLIP Score of original and counterfactual images in Causal-HalBench with the text prompt ``a photo of \{object\}". Here, {object} includes both the target object and the counterfactual object.}
    \label{tab:CLIP Score}
\end{table}

\section{Conclusion}

Our study first introduces causal analysis to LVLM hallucination research. We propose that during training, models learn co-occurrence bias between highly co-occurring objects, leading to misjudgments in object recognition. This phenomenon is termed spurious correlation. To evaluate this, we create a new benchmark called Causal-HalBench. It uses advanced proprietary LVLMs and inpainting models to construct counterfactual samples, thereby breaking spurious correlations in images, and introduces new causality-based metrics to quantify the impact of these spurious correlations. Experiments show that all current models are affected by this issue to various degree, highlighting a key way to make them more accurate in the future.

\section*{Acknowledgments}
This research is supported by the National Science and Technology Major Project (2023ZD0121102).

\bibliography{aaai2026}

\end{document}